\documentclass[twoside,11pt]{article}

\usepackage{jmlr2e}
\usepackage{url}

\usepackage{amsbsy}
\usepackage{amsmath}
\usepackage{mathrsfs}
\usepackage{dsfont}
\usepackage{hyperref}
\usepackage{tikz}
\usetikzlibrary{shapes.geometric}
\usepackage{caption}
\usepackage{subfig}
\usepackage{makecell}
\usepackage{verbatim}

\newtheorem{axsy}[theorem]{Axiomatic System}

\newenvironment{proofx}{\par\noindent{\bf Proof\ }}

\newcommand{\upceil}[1]{\lceil {#1} \rceil}
\newcommand{\lrangle}[1]{\langle {#1} \rangle}

\newcommand{\subsym}[1]{\mathfrak{{#1}}}
\newcommand{\newdefsymbol}{$\bigstar$}

\newcommand{\HX}{H_X}

\newcommand{\LX}{\mathscr{L}_X}
\newcommand{\n}{\neg}
\newcommand{\infer}{\rightarrow}

\newcommand{\Tone}{\mathfrak{T}}
\newcommand{\Ttwo}{\mathfrak{T}'}

\newcommand{\Godel}{G{\"o}del}

\newcommand{\IOeq}{\overset{IO}{=}}
\newcommand{\TM}{\mathcal{TM}}
\newcommand{\manyone}{\leq_m}
\newcommand{\compleit}[1]{\overline{{#1}}}
\newcommand{\HALT}{Halt}
\newcommand{\HALTc}{\compleit{Halt}}
\newcommand{\SAMEIO}{Same}
\newcommand{\SAMEIOc}{\compleit{Same}}
\newcommand{\DesiredOne}{DesiredOne}

\firstpageno{1}

\begin{document}

\title{G{\"o}del's Sentence Is An Adversarial Example But Unsolvable}


\author{\name Xiaodong Qi \email qixiaodong@hust.edu.cn \\
       \name Lansheng Han* \email hanlansheng@hust.edu.cn \\
       \addr School of Cyber Science and Engineering \\
       Huazhong University of Science and Technology\\
       Wuhan, Hubei, China}

\editor{}

\maketitle

\begin{abstract}
  In recent years, different types of adversarial examples from different fields have emerged endlessly, including purely natural ones without perturbations. A variety of defenses are proposed and then broken quickly. Two fundamental questions need to be asked: What's the reason for the existence of adversarial examples and are adversarial examples unsolvable?
  In this paper, we will show the reason for the existence of adversarial examples is there are non-isomorphic natural explanations that can all explain data set.
  Specifically, for two natural explanations of being true and provable, G{\"o}del's sentence is an adversarial example but ineliminable. It can't be solved by the re-accumulation of data set or the re-improvement of learning algorithm.
  Finally, from the perspective of computability, we will prove the incomputability for adversarial examples, which are unrecognizable.

\end{abstract}


\section{Introduce}
Adversarial examples have attracted significant attention in recent years. They have emerged in image\citep{goodfellow2014explaining,engstrom2017rotation}, audio\citep{carlini2018audio,taori2019targeted,qin2019imperceptible}, text classification\citep{samanta2017towards,lei2019discrete} and NLP \citep{alzantot2018generating,niven2019probing,zhang2019generating,eger2019text}.
The reason for their existence remains unclear. Previous work has proposed different answers from the perspective of data set and learning algorithm, including the data set is not big enough\citep{schmidt2018adversarially}, not good enough\citep{ford2019adversarial} or not random enough\citep{weiguang2019sensitivity}, and the learning algorithm is not complex enough\citep{bubeck2018adversarial,nakkiran2019adversarial} or not robust enough\citep{xiao2018training,stutz2019disentangling}.
Moreover, in \citet{shafahi2018adversarial}, it's shown that for certain classes of problems (on the sphere or cube), adversarial examples are inescapable. In this paper, for a simple binary classification problem, we will propose an adversarial example---\emph{\Godel{}'s sentence}, which is not caused by data set or learning algorithm. In our example, for any finite or any decidable infinite data set, for any learning algorithm, \Godel{}'s sentence will always be there as an adversarial example and ineliminable. It can't be solved by the re-accumulation of data set or the re-improvement of learning algorithm.

First of all, what are adversarial examples? The definition of adversarial example now is almost all based on perturbations. For instance, in \citet{shafahi2018adversarial}, a formal definition of an adversarial example is as follows.

\emph{Consider a point $x \in \Omega$ drawn from class $c$; a scalar $\epsilon > 0$; and a metric $d$. We say that $x$ admits an $\epsilon$-adversarial example in the metric $d$ if there exists a point $\hat{x} \in \Omega$ with $\mathcal{C}(\hat{x}) \neq c$; and $d(x,\hat{x}) \leq \epsilon$.} ($\mathcal{C}:\Omega \to \{1,2,...,m\}$ is a ``classifier'' function.)

\begin{figure}[htb!]
  \footnotesize
  \centering
  \begin{tabular}{ccc}
  \includegraphics[width=.25\textwidth]{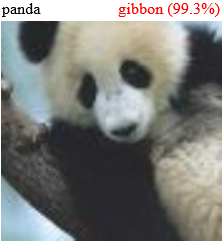} & \includegraphics[width=.29\textwidth]{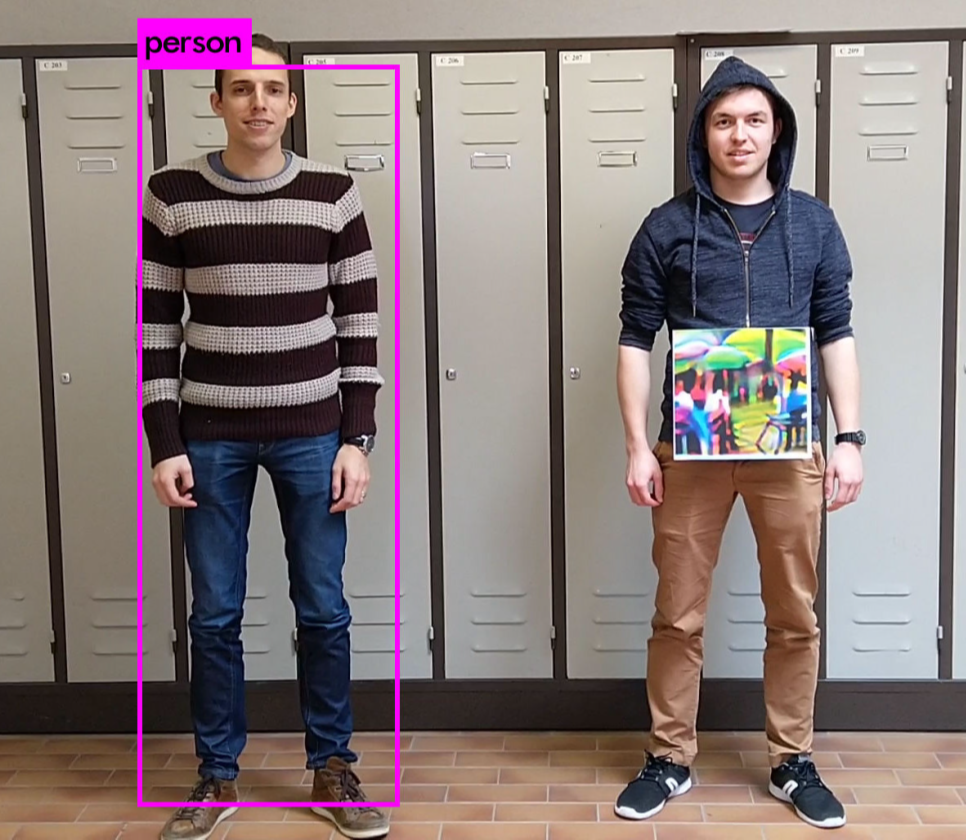} & \includegraphics[width=.25\textwidth]{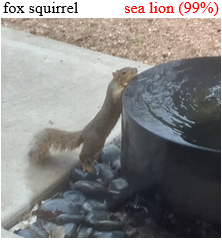} \\
  \makecell*[l]{(a) Perturbations in digital\\ \ \ \ \ \ world.} &\makecell*[l]{(b) Perturbations in real world.\\} &\makecell*[l]{(c) Without perturbations,\\ \ \ \ \ \ natural example.}\\
  \end{tabular}
  \caption{Three different types of adversarial examples~\citep{goodfellow2014explaining,thys2019fooling,hendrycks2019natural} }
  \label{figure:threetypes}
\end{figure}

However, adversarial examples are not limit to the type of perturbations, like Figure~\ref{figure:threetypes}.a \& \ref{figure:threetypes}.b\citep{goodfellow2014explaining,thys2019fooling}. Recent work has shown there are many purely natural adversarial examples without any perturbations, like Figure~\ref{figure:threetypes}.c\citep{hendrycks2019natural}.
It's inappropriate to understand the latter as the normal wrong outputs and only the former as the adversarial examples.
Otherwise we will fall into an infinite regress to argue which special type of wrong output is an adversarial example and which is not, because any insular definition will be so difficult to reconcile with the facts observed in practice. Therefore, it's necessary to redefine adversarial example beyond perturbations, which is also important for us to apprehend the reason for their existence, and their unsolvability in our example.

Second, are adversarial examples unsolvable? In this paper, the specific adversarial example, \Godel{}'s sentence, is ineliminable.
It can't be solved by re-accumulation of data set. Adversarial training\citep{goodfellow2014explaining,wu2017adversarial} can't be helpful because of the ineradicable incompleteness.
It can't be solved by re-improvement of learning algorithm, either.
Each learning algorithm must be equivalent to a Turing machine, no matter it's based on logistic regression, SVM or neural network. However, changing one Turing machine to another Turing machine won't be helpful, because no Turing machine can learn something from nothing.

Finally, before we redefine adversarial example and study their unsolvability, some essential issues about them in machine learning need to be clarified.
Especially, what is the ground truth, what is to be learned, what can be expressed by data set, and what is expected to express by us through data set? Maybe the answers to these issues are so self-evident that they are ignored for a long time. Actually, none of these answers is clear, although they are thought self-evident. However, Russell's paradox always reminds us that danger is always in the unclarity. Once these issues are clarified, the reason for the existence of adversarial examples will be clear.
There are three main contributions in this paper.
\begin{enumerate}\setlength{\itemsep}{0.2ex}
  \item We will point out the reason for the existence of adversarial examples is there are non-isomorphic natural explanations that can all explain data set.
  \item We will show that G{\"o}del's sentence is an adversarial example but ineliminable.
  \item We will prove the incomputability for adversarial examples, which are unrecognizable.
\end{enumerate}

\section{Related Work}
Adversarial examples are first demonstrated in \citet{szegedy2013intriguing} and \citet{biggio2013evasion} in digital world. They can be generated by adding pixel-level changes to a normal image\citep{goodfellow2014explaining,papernot2017practical} or doing simple rotation and translation to a normal image\citep{engstrom2017rotation,alcorn2019strike}.
They also exist in real world. By placing a few stickers in real world, a physical stop sign is recognized as a speed limit 45 sign\citep{eykholt2018robust}, images are misclassified\citep{liu2019perceptual}, and Autopilot goes to the wrong lane, because it identifies stickers as road traffic markings\citep{TeslaAutopilotThreeSmallStickersIEEE26}.
By wearing a adversarial patch in real world, the person is difficult to be identified\citep{thys2019fooling,komkov2019advhat}.
There are also adversarial examples of audio in real world\citep{li2019adversarial}.
Excluding the perturbations in digital world (pixel-level changes, rotation and translation) and in real world (stickers and patches), recent work has shown there have been plenty of purely natural adversarial examples in real world without any perturbations \citep{hendrycks2019natural}, like Figure~\ref{figure:threetypes}.c.

Many solutions for adversarial examples have been proposed, like adversarial training\citep{goodfellow2014explaining,wu2017adversarial}, network distillation\citep{papernot2016distillation}, classifier robustifying\citep{abbasi2017robustness}, adversarial detecting\citep{lu2017safetynet}, input reconstruction\citep{gu2014towards} and network verification\citep{katz2017reluplex}. However, almost all solutions are shown to be effective only for part of adversarial attacks \citep{yuan2019adversarial}.

Recent work has also made encouraging progress to clarify some essential issues about adversarial examples.
What can be expressed by data set? More technically, what is the meaning of label in data set?
The extraordinary work in \citet{ghorbani2019towards} shows the difference between ``$basketball$'' and basketball, the former is a meaningless label in data set, and the latter is a word meaning a sphere or a sport.
It's shown that the most important concept for predicting ``$basketball$'' images is the players' jerseys rather than the ball itself. Moreover, even if there are only the basketball jerseys without ball itself in a image, it can still be classified as the class ``$basketball$''.
What is to be learned? The remarkable work in \citet{ilyas2019adversarial} shows that what are learned by machine can be non-robust features, but what human can understand and will want machine to learn are robust features, however they are both in data set, and adversarial example is a natural consequence of the presence of highly predictive but non-robust features in standard data sets.


\section{Meaning of Label}

To interpret what can be expressed by data set and reveal that it's quite different from what is expected to express by us through data set, we need to figure out what the label in data set means, which is crucial for us to understand why there can be different explanations that can all explain data set. Here is a binary classification problem for cats and dogs.

\subsection{Where Are Cats?}

\begin{figure}[htb!]
  \centering
  \begin{tabular}{m{1.5cm}m{0.8cm}m{0.8cm}m{0.8cm}m{0.8cm}m{0.2cm}m{1.5cm}m{0.8cm}m{0.8cm}m{0.8cm}m{0.8cm}}
   &\multicolumn{4}{c}{Label} && &\multicolumn{4}{c}{Label} \\
  Image-\uppercase\expandafter{\romannumeral1} &type1 &type2 &type3 &type4 &&Image-\uppercase\expandafter{\romannumeral2} &type1 &type2 &type3 &type4\\ \\
  \includegraphics[width=.08\textwidth]{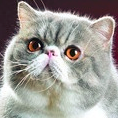} &\makecell[c]{$cat$} &\makecell[c]{$chat$} &\makecell[c]{$0$} &\makecell[c]{\underline{$dog$}} &&\includegraphics[width=.08\textwidth]{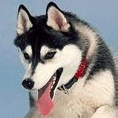} &\makecell[c]{$dog$} &\makecell[c]{$chien$} &\makecell[c]{$1$} &\makecell[c]{\underline{$cat$}}\\
  \includegraphics[width=.08\textwidth]{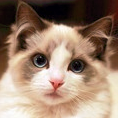} &\makecell[c]{$cat$} &\makecell[c]{$chat$} &\makecell[c]{$0$} &\makecell[c]{\underline{$dog$}} &&\includegraphics[width=.08\textwidth]{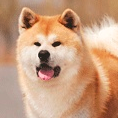} &\makecell[c]{$dog$} &\makecell[c]{$chien$} &\makecell[c]{$1$} &\makecell[c]{\underline{$cat$}}\\
  \includegraphics[width=.08\textwidth]{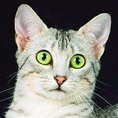} &\makecell[c]{$cat$} &\makecell[c]{$chat$} &\makecell[c]{$0$} &\makecell[c]{\underline{$dog$}} &&\includegraphics[width=.08\textwidth]{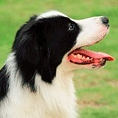} &\makecell[c]{$dog$} &\makecell[c]{$chien$} &\makecell[c]{$1$} &\makecell[c]{\underline{$cat$}}\\
  \end{tabular}
  \caption{Data set of cats and dogs}
  \label{figure:animals}
\end{figure}

The data set is in Figure~\ref{figure:animals}. If we were native English speakers, we would adopt the label of type1. And, where are cats?
\vspace{-0.2cm}

\begin{table}[h]\footnotesize
\centering
\begin{tabular}{|p{1cm}|p{5.6cm}|p{1cm}|p{5.9cm}|}
\hline
\multicolumn{2}{|c|}{\small{\textbf{For Human}}} & \multicolumn{2}{c|}{\small{\textbf{For Machine}}}\\
\hline
Fact.$1_h$  &\makecell*[l]{In Image-\uppercase\expandafter{\romannumeral1}, there are images of animals\\ who can meow and catch mice.} &Fact.$1_m$ &\makecell*[l]{In Image-\uppercase\expandafter{\romannumeral1}, there are some legal png files.} \\
\hline
Fact.$2_h$  &\makecell*[l]{The label of left type1 is ``$cat$'', a word\\ which means the animal who can meow\\ and catch mice.} &Fact.$2_m$ &\makecell*[l]{The label of left type1 is ``$cat$'', a symbolic\\ string which is meaningless.} \\
\hline
Fact.$3_h$  &\makecell*[l]{Combining the Image-\uppercase\expandafter{\romannumeral1} and the label of\\ left type1, it means that the animals in\\ Image-\uppercase\expandafter{\romannumeral1} are all cats, the animals who\\ can meow and catch mice.} &Fact.$3_m$ &\makecell*[l]{Combining the Image-\uppercase\expandafter{\romannumeral1} and the label of\\ left type1, it means that the images in\\ Image-\uppercase\expandafter{\romannumeral1} are all of the same type, denoted\\ as ``$cat$''.} \\
\hline
\end{tabular}
\caption{Understandings of data set for human and machine are different}
\label{table:fact}
\end{table}
\vspace{-0.2cm}

%

For machine, there is no cat, the animal who can meow and catch mice; there is only ``$cat$'', the meaningless symbolic string (Table~\ref{table:fact}). There are more reasons (see Figure~\ref{figure:animals}). In French, there is no cat but $chat$. When programming, there is no cat but $0$. In a planet in the Andromeda Galaxy with the isomorphic civilization of human's, there is no cat but \underline{$dog$}. The only meaning of a label is to show that those with the same label belong to the same class, and those with different labels belong to different classes. As for the symbolic representation of a label, it doesn't matter.

\subsection{Can Smart Enough Algorithm Find Cats?}

To answer this question, we need to detail the  current machine learning paradigm. All we have are the data set and learning algorithm. The data set is divided into two (or three) parts, one of them is used to train by learning algorithm, another one of them is used to test. However, there is no cat in the data set, and nobody can learn something from nothing. The good performance on test set means it performs well to distinguish ``$cat$'' from ``$dog$''. But the ``$cat$'' and ``$dog$'' is not the same thing as the cat and dog we know, the former can meow and the latter can bark. For instance, as a well-known example of algorithmic bias, Google identifies black people as gorillas~\citep{crawford2016artificial}. However, the truth is that, Google identifies black people as ``$gorillas$'', the meaningless symbolic string, and we interpret the ``$gorillas$'' as gorillas we know. One might wonder, how could ``$gorillas$'' not be interpret as gorillas we know?
However, we are not playing with words, it's just that some basic concepts in machine learning have not been established.

\section{Reason for Existence of Adversarial Examples}

The symbolic representation of a label doesn't matter, but the relationship between labels matters.
For example, the meaning of ``$cat$'' and ``$dog$'' can't be expressed by the data set in Figure~\ref{figure:animals}, but the following three points can be expressed by the data set: (1) the images in Image-\uppercase\expandafter{\romannumeral1} are of the same type; (2) the images in Image-\uppercase\expandafter{\romannumeral2} are of the same type; (3) these two types are not the same.
We need to depict the partial isomorphism in a more precise way,
then define adversarial example.

\begin{definition}
For any field {\small$\varPi$}, let its universal set be {\small$U_\varPi$}, for any {\small$Y \neq \varnothing$}, any function {\small$f:U_\varPi \rightarrow Y$} is called an {\large\emph{explanation}} of field {\small$\varPi$}.
\end{definition}

\begin{definition}
Let {\small$f:U_\varPi \rightarrow Y$} be an explanation of field {\small$\varPi$}, {\small$X_\varPi \subseteq U_\varPi$}, {\small$f(X_\varPi)=\{f(x) \mid x \in X_\varPi\}$}, the surjection {\small$f^{X_\varPi}:X_\varPi \rightarrow f(X_\varPi)$} satisfying that {\small$\forall x \in X_\varPi,f^{X_\varPi}(x)=f(x)$} is called the {\large\emph{explanation of {\small$f$} limited to {\small$X_\varPi$}}}.
\end{definition}

\begin{definition}
Let {\small$f_1:U_\varPi \rightarrow Y_1$} and {\small$f_2:U_\varPi \rightarrow Y_2$} be explanations of field {\small$\varPi$}, let {\small$X_\varPi \subseteq U_\varPi$}, if there is a bijection {\small$g$} between {\small$f_1(X_\varPi)$} and {\small$f_2(X_\varPi)$} that {\small$\forall x \in X_\varPi, f_1^{X_\varPi}(x)=y_1 \Leftrightarrow f_2^{X_\varPi}(x)=y_2 \wedge y_2=g(y_1)$}, it's called that {\large\emph{the explanations {\small$f_1$} and {\small$f_2$} are isomorphic on {\small$X_\varPi$}}}. If the explanations {\small$f_1$} and {\small$f_2$} are isomorphic on {\small$U_\varPi$}, it's simply called that the {\large\emph{explanations {\small$f_1$} and {\small$f_2$} are isomorphic}}.
\end{definition}

\begin{definition}
Let {\small$f:U_\varPi \rightarrow Y$} be an explanation of field {\small$\varPi$} and {\small$X_\varPi \subseteq U_\varPi$}, {\small$\{(x,f^{X_\varPi}(x)) \mid x \in X_\varPi\}$} is denoted as {\small$\upceil{X_\varPi,f^{X_\varPi}(X_\varPi)}$}, if {\small$h:U_\varPi \rightarrow Z$} is an explanation of field {\small$\varPi$}, and the explanations {\small$f$} and {\small$h$} are isomorphic on {\small$X_\varPi$}, it's called {\small$h$} {\large\emph{explains {\small$\upceil{X_\varPi,f^{X_\varPi}(X_\varPi)}$}}}, also called {\small$h$} is an {\large\emph{explanation on {\small$\upceil{X_\varPi,f^{X_\varPi}(X_\varPi)}$}}}.
\end{definition}

\begin{definition}
Let {\small$f:U_\varPi \rightarrow Y$} be an explanation of field {\small$\varPi$}, {\small$X_\varPi \subseteq U_\varPi$}, {\small$h_1,h_2,h_3,\cdots,h_n$} are the enumeration of all the non-isomorphic {\large\emph{natural explanations}} on {\small$\upceil{X_\varPi,f^{X_\varPi}(X_\varPi)}$}. If there is a set {\small$\mathds{G}(\upceil{X_\varPi,f^{X_\varPi}(X_\varPi)}) \subseteq U_\varPi$}, satisfying that {\small$X_\varPi \subseteq \mathds{G}(\upceil{X_\varPi,f^{X_\varPi}(X_\varPi)})$}, and for any {\small$i,j \leq n \wedge i \neq j$}, {\small$h_i$} and {\small$h_j$} are isomorphic on {\small$\mathds{G}(\upceil{X_\varPi,f^{X_\varPi}(X_\varPi)})$}, and for any {\small$\mathds{G}(\upceil{X_\varPi,f^{X_\varPi}(X_\varPi)}) \subsetneq W \subseteq U_\varPi$}, there are {\small$i,j \leq n \wedge i \neq j$} to make that {\small$h_i$} and {\small$h_j$} are not isomorphic on {\small$W$}, {\small$\mathds{G}(\upceil{X_\varPi,f^{X_\varPi}(X_\varPi)})$} is called a {\large\emph{generalization set of {\small$\upceil{X_\varPi,f^{X_\varPi}(X_\varPi)}$}}}.
\end{definition}

\begin{definition}
Let {\small$f:U_\varPi \rightarrow Y$} be an explanation of field {\small$\varPi$}, {\small$X_\varPi \subseteq U_\varPi$}, if {\small$\mathds{G}(\upceil{X_\varPi,f^{X_\varPi}(X_\varPi)})$} is the generalization set of {\small$\upceil{X_\varPi,f^{X_\varPi}(X_\varPi)}$}, {\small$\mathds{A}(\upceil{X_\varPi,f^{X_\varPi}(X_\varPi)})=U_\varPi - \mathds{G}(\upceil{X_\varPi,f^{X_\varPi}(X_\varPi)})$}, {\small$\mathds{A}(\upceil{X_\varPi,f^{X_\varPi}(X_\varPi)})$} is called the {\large\emph{adversarial set of {\small$\upceil{X_\varPi,f^{X_\varPi}(X_\varPi)}$}}}.
\end{definition}

\begin{table}[h]\small
\centering
\begin{tabular}{|p{2.8cm}|p{2cm}|p{7.6cm}|}
\hline
\multicolumn{1}{|c|}{Name} &\multicolumn{1}{c|}{Symbol} &\multicolumn{1}{c|}{Description}\\
\hline
Object Set &\multicolumn{1}{l|}{$U_\varPi$} & the universal set of field $\varPi$.\\
\hline
Known Set &\multicolumn{1}{l|}{$X_\varPi$} & $X_\varPi \subseteq U_\varPi$, the set known to us.\\
\hline
Unknown Set &\multicolumn{1}{l|}{$U_\varPi-X_\varPi$} & the set unknown to us.\\
\hline
Training Set & & $X_\varPi^1,X_\varPi^2,X_\varPi^3 \subseteq X_\varPi$,\\
Verification Set &\multicolumn{1}{l|}{$X_\varPi^1,X_\varPi^2,X_\varPi^3$} & $X_\varPi^1 \cup X_\varPi^2 \cup X_\varPi^3 = X_\varPi$,\\
Test Set & & $X_\varPi^i \cap X_\varPi^j = \varnothing, \forall i \neq j, i,j=1,2,3$.\\
\hline
Data Set &\multicolumn{1}{l|}{$X_\varPi/\upceil{X_\varPi,f^{X_\varPi}(X_\varPi)}$} & without/with explanatory information.\\
\hline
Generalization Set &\multicolumn{1}{l|}{$\mathds{G}(\upceil{X_\varPi,f^{X_\varPi}(X_\varPi)})$} & $X_\varPi \subseteq \mathds{G}(\upceil{X_\varPi,f^{X_\varPi}(X_\varPi)}) \subseteq U_\varPi$.\\
\hline
Adversarial Set &\multicolumn{1}{l|}{$\mathds{A}(\upceil{X_\varPi,f^{X_\varPi}(X_\varPi)})$} & $\mathds{A}(\upceil{X_\varPi,f^{X_\varPi}(X_\varPi)}) = U_\varPi - \mathds{G}(\upceil{X_\varPi,f^{X_\varPi}(X_\varPi)})$.\\
\hline
\end{tabular}
\caption{\textbf{Definition of sets} Training set, verification set and test set are the subsets of known set. Known set is a subset of object set. Generalization set and adversarial set depend entirely on known set (with explanatory information) and have nothing to do with learning algorithms. What is known to us is the known set, but what people pay attention to is always the object set.}
\label{table:set}
\end{table}

\begin{figure}[htb!]\footnotesize
\centering
\begin{tikzpicture}
    [L1Node/.style={circle, draw=black!100, fill=black!100, minimum size=30mm},
    L2Node/.style={circle, draw=black!50, fill=black!50, minimum size=20mm},
    L3Node/.style={circle, draw=black!10, fill=black!10, minimum size=10mm},
    L4Node/.style={circle, draw=black!100, fill=black!100, minimum size=6mm},
    L5Node/.style={circle, draw=black!50, fill=black!50, minimum size=6mm},
    L6Node/.style={circle, draw=black!10, fill=black!10, minimum size=6mm},
    L7Node/.style={circle, draw=black!50, fill=black!50, minimum size=4mm},
    L8Node/.style={circle, draw=black!10, fill=black!10, minimum size=3.2mm}]
       \node[L4Node,label=left:{\footnotesize object set}] (n7) at (1, 11){};
       \node[L5Node,label=left:{\footnotesize generalization set}] (n8) at (1, 10){};
       \node[L6Node,label=left:{\footnotesize known set}] (n9) at (1, 9){};
       \node[L4Node,label=right:{\footnotesize adversarial set}] (n11) at (7, 11){};
       \node[L4Node,label=right:{\footnotesize unknown set}] (n12) at (7, 10){};
       \node[L7Node] (n13) at (7, 10.9){};
       \node[L8Node] (n14) at (7, 9.86){};
       \node[L6Node,label=right:{\begin{tabular}{l}
                                 {\footnotesize training set}\\
                                 {\footnotesize verification set}\\
                                 {\footnotesize test set}
                                 \end{tabular}}] (n10) at (7,9){};
       \draw[] (7,9) -- (7,9.3);
       \draw[] (6.7,9) -- (7.3,9);
       \node[L1Node] (n4) at (4, 10){};
       \node[L2Node] (n5) at (4, 9.5){};
       \node[L3Node] (n6) at (4, 9){};
       \draw[] (4,9) -- (4,9.5);
       \draw[] (3.5,9) -- (4.5,9);
\end{tikzpicture}
\caption{\textbf{Relationship between sets} Known set $\subseteq$ generalization set $\subseteq$ object set. To keep expanding known set, generalization set will inevitably expand together, but it doesn't mean adversarial set will necessarily be empty gradually, like the example in Section~\ref{section:GodelSentence}.}
\label{figure:set}
\end{figure}
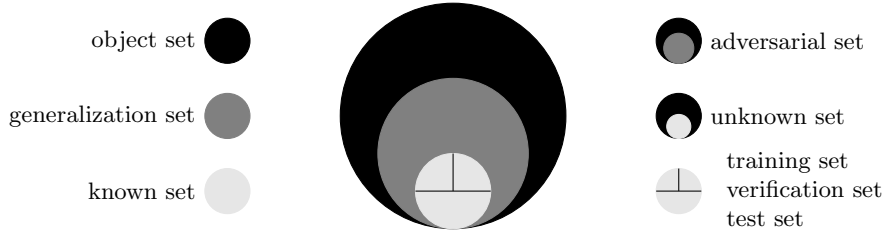

For explanations, they are further divided into {\large\emph{natural explanations}} and {\large\emph{unnatural explanations}}. The former are the explanations that conform to human cognition of reality, and the latter are the explanations that don't.
There are always many unnatural explanations that can explain the data set, but they are not convincing.
What we concern are always natural explanations,
but there may be non-isomorphic natural explanations that can all explain the data set.
For example, the {\large\emph{classical mechanics}} (from Newton) and {\large\emph{special relativity}} (from Einstein), the {\large\emph{Euclidean}} and {\large\emph{Lobachevskian}} geometry, the {\large\emph{ZF}} and {\large\emph{ZFC}} in axiomatic set theory, they are all non-isomorphic natural explanations. They can be isomorphic respectively on the the ground truth we have observed in earth before 1887, the absolute geometry, and $ZF$.
Based on the natural explanations, the generalization set and adversarial set can be defined (Table~\ref{table:set}, Figure~\ref{figure:set}). The adversarial example can be defined in this way.

\begin{definition} \label{defn:adversarialexample}
For any $\upceil{X_\varPi,f^{X_\varPi}(X_\varPi)}$, $\forall e_\varPi \in U_\varPi$, if $e_\varPi \in \mathds{A}(\upceil{X_\varPi,f^{X_\varPi}(X_\varPi)})$, $e_\varPi$ is an \textbf{adversarial example} of $\upceil{X_\varPi,f^{X_\varPi}(X_\varPi)}$.
\end{definition}

The reason for the existence of adversarial examples is that there are non-isomorphic natural explanations that can all explain data set. Any $e_\varPi$ in adversarial set is an adversarial example, because it's unknown which explanation is the desired one, but the outputs of non-isomorphic natural explanations (on the data set) are different, so any output can be wrong.
All in our mind is the desired explanation we want to express, and according to this explanation, we generate lots of data and make up the data set. However, we may not notice that there is an undesired natural explanation on the data set.
If the outcome learned by the learning algorithm happens to be the undesired one, it can perform perfectly on the known set $X_\varPi$, including the training set $X_\varPi^1$ and test set $X_\varPi^3$, but the output for the adversarial example will be considered to be wrong by us.

\section{Incompleteness and \Godel{}'s Sentence}

In this section, we are concerning about two very specific natural explanations, {\large\emph{being true}} and {\large\emph{being provable}}. Meanwhile, we will show there are indeed two non-isomorphic natural explanations on a million-scale data set. Moreover, the two natural explanations are not insubstantial like castle in the air. They are quite comprehensible and can both be implemented by concrete and simple algorithms. We will show whether being true and provable are isomorphic depends on the completeness theorem. For simplicity, let's start with propositional logic.

\subsection{Peirce's Law} \label{section:Peirce'sLaw}

Here is a data set for a binary classification problem in Table~\ref{table:HX}, and the full data set can be found here\footnote{https://github.com/IcePulverizer/Hx}, where there are millions of formulas for each class.
It's easy to see the formulas in class $\mathbb{T}$ are all tautologies, and formulas in class $\mathbb{C}$ are all contradictions. We can tell the readers that this assertion is still true for the full data set. However, after machine learning based on this data set, which class should the Peirce's law be classified into?
\begin{center}
Peirce's law: $((p \infer q) \infer p) \infer p$.
\end{center}

\begin{table}[h] \small
\centering
\begin{tabular}{|p{7cm}|p{7cm}|}
\hline
\multicolumn{1}{|c|}{Class $\mathbb{T}$} & \multicolumn{1}{c|}{Class $\mathbb{C}$}\\
\hline
$p \infer (q \infer p)$ & $\n(p \infer (q \infer p))$ \\
\hline
$(p \infer (q \infer r)) \infer ((p \infer q) \infer (p \infer r))$ & $\n((p \infer (q \infer r)) \infer ((p \infer q) \infer (p \infer r)))$ \\
\hline
$\n p \infer (p \infer q)$ & $\n(\n p \infer (p \infer q))$ \\
\hline
$(p \infer \n p) \infer \n p$ & $\n((p \infer \n p) \infer \n p)$ \\
\hline
$p \infer \n \n p$ & $\n(p \infer \n \n p)$ \\
\hline
$\n \n \n \n p \infer \n \n p$ & $\n(\n \n \n \n p \infer \n \n p)$ \\
\hline
$\n \n (\n \n \n \n p \infer \n \n p)$ & $\n \n \n(\n \n \n \n p \infer \n \n p)$ \\
\hline
$(p \infer q) \infer (\n q \infer \n p)$ & $\n((p \infer q) \infer (\n q \infer \n p))$ \\
\hline
$(p \infer \n q) \infer (q \infer \n p)$ & $\n((p \infer \n q) \infer (q \infer \n p))$ \\
\hline
$\n ((\n p \infer \n p) \infer \n (q \infer \n \n q))$ & $(\n p \infer \n p) \infer \n (q \infer \n \n q)$ \\
\hline
$\n \n (((p \infer q) \infer p) \infer p)$ & $\n(((p \infer q) \infer p) \infer p)$ \\
\hline
\end{tabular}
\caption{Data set of formulas}
\label{table:HX}
\end{table}

\begin{minipage}{\textwidth}
\begin{minipage}[t]{0.45\textwidth}
\centering
\makeatletter\def\@captype{table}\makeatother
\begin{tabular}{c|c}
 & $f_\n$ \\
\hline
$0$ & $1$ \\
$1$ & $0$ \\
\multicolumn{2}{c}{}
\end{tabular}
\begin{tabular}{cc}
 &  \\
 &  \\
\end{tabular}
\begin{tabular}{c|cc}
$f_\infer$ & $0$ & $1$ \\
\hline
$0$ & $1$ & $1$ \\
$1$ & $0$ & $1$ \\
\multicolumn{2}{c}{}
\end{tabular}
\caption{$\Tone=\lrangle{\{0,\textbf{1}\},f_\n,f_\infer}\mid$true}
\label{table:Tone}
\end{minipage}
\begin{minipage}[t]{0.45\textwidth}
\centering
\makeatletter\def\@captype{table}\makeatother
\begin{tabular}{c|c}
 & $g_\n$ \\
\hline
$0$ & $1$ \\
$1$ & $2$ \\
$2$ & $1$
\end{tabular}
\begin{tabular}{cc}
 &  \\
 &  \\
 &  \\
 &
\end{tabular}
\begin{tabular}{c|ccc}
$g_\infer$ & $0$ & $1$ & $2$\\
\hline
$0$ & $2$ & $1$ & $2$\\
$1$ & $2$ & $2$ & $2$\\
$2$ & $0$ & $1$ & $2$
\end{tabular}
\caption{$\Ttwo=\lrangle{\{0,1,\textbf{2}\},g_\n,g_\infer}\mid$provable}
\label{table:Ttwo}
\end{minipage}
\end{minipage}
\vspace{-0.0cm}

It depends on how to explain class $\mathbb{T}$ and $\mathbb{C}$. There are at least two natural explanations. The apparent one is {\large\emph{being true}}, which can be explained by the two-value interpretation structure $\Tone$ in Table~\ref{table:Tone}, and the formulas in class $\mathbb{T}$ are $\Tone$-$\{1\}$-$tautologies$. Another one is {\large\emph{being provable}} in $\HX$, which can explained by the three-value interpretation structure $\Ttwo$ in Table~\ref{table:Ttwo}, and the formulas in class $\mathbb{T}$ are $\Ttwo$-$\{2\}$-$tautologies$. $\HX$ is an axiomatic system in formal language $\LX$. The axioms and rules of inference of $\HX$ are in Axiomatic System~\ref{expl:HX}.
The Peirce's law is true, but unprovable in $\HX$. According to $\Tone$, the Peirce's law should be classified into class $\mathbb{T}$ because it's true. However, according to $\Ttwo$, the Peirce's law should be classified into class $\mathbb{C}$ because it's unprovable in $\HX$.

\begin{axsy} \label{expl:HX}
$\HX$.

The axioms of $\HX$:
\begin{itemize}\setlength{\itemindent}{1em}
  \item[\emph{X1}] $p \infer (q \infer p)$
  \item[\emph{X2}] $(p \infer (q \infer r)) \infer ((p \infer q) \infer (p \infer r))$
  \item[\emph{X3}] $\n p \infer (p \infer q)$
  \item[\emph{X4}] $(p \infer \n p) \infer \n p$
\end{itemize}

The rules of inference of $\HX$:
\begin{itemize}\setlength{\itemindent}{1em}
  \item[mp] (modus ponens): $\psi$ can be obtained from $\phi \infer \psi$ and $\phi$.
  \item[sub] 	(substitution): $\phi(\subsym{s})$ can be obtained from $\phi$, where $\subsym{s}$ is a finite substitution. \hfill\BlackBox
\end{itemize}
\end{axsy}

%
%

\begin{theorem}[Soundness] \label{thm:HXSoundness}
In $\HX$, if $\phi$ is provable, $\phi$ is a tautology.
\end{theorem}

\begin{theorem}[Incompleteness] \label{thm:HXIncompleteness}
In $\HX$, if $\phi$ is a tautology but not a $\Ttwo$-$\{2\}$-$tautology$, $\phi$ is unprovable.
\end{theorem}

By this instance, vivid answers are given to two questions that what is the ground truth and what is to be learned. They are both subjective and rely on which explanation is desired by us, and the data set avails nothing. Furthermore, as the builders of the data set in Table~\ref{table:HX}, even if we admit what we want to express by data set is not the formula that is a tautology or contradiction but the $\LX$-formula that is provable or unprovable in $\HX$, does this admission matter? We can't deny that even though the data set is generated by us according to whether a $\LX$-formula is provable or unprovable in $\HX$ by $\Ttwo$, it can also be explained by whether a formula is a tautology or contradiction by $\Tone$. The idea of the builders of the data set doesn't matter at all, which is important for us to understand adversarial examples.

In our data set, the Peirce's law is an adversarial example, because there are two non-isomorphic natural explanations that can both explain the data set. One of them is {\large\emph{being true}} by $\Tone$. Another one is {\large\emph{being provable}} (in $\HX$) by $\Ttwo$. They two are non-isomorphic because the completeness theorem doesn't hold in $\HX$. The {\large\emph{soundness theorem}} guarantees that anything provable must be true (Theorem~\ref{thm:HXSoundness}). The {\large\emph{completeness theorem}} guarantees that anything true must be provable. Hence, whether  being true and provable are isomorphic depends on the soundness and completeness theorem. Since no one can stand an unsound axiomatic system, we only need to concern about the completeness. However, $\HX$ is incomplete (Theorem~\ref{thm:HXIncompleteness}). The Peirce's law is true, but unprovable in $\HX$.

\subsection{\Godel{}'s Sentence}\label{section:GodelSentence}


\begin{theorem}[\Godel{}'s Incompleteness] \label{thm:godel}
Let $T \ (PA \subseteq T)$ be an axiomatic system, if $T$ is consistent, there is a sentence $\sigma_T$ so that $\sigma_T$ and $\n \sigma_T$ are both unprovable in T.
\end{theorem}

Beyond propositional logic, the \Godel{}'s sentence $\sigma_T$, in any consistent first-order axiomatic system $T$ that is sufficient to contain all axioms of $PA$ (first-order Peano arithmetic axiomatic system), is also an adversarial example. \Godel{}'s sentence is constructed by~\citet{godel1931formal}, then improved by~\citet{rosser1936extensions}, it's true but unprovable inside system. Theorem~\ref{thm:godel} is famous as {\large\emph{\Godel's first incomplete theorem}}. \Godel{}'s sentence fits our definition of adversarial example, perfectly. First, no one can deny that being true and provable are two natural explanations. Second, for any consistent existing data set, there must be an axiomatic system $T$ to guarantee being true and provable are isomorphic on this data set. Third, there will be the \Godel{}'s sentence $\sigma_T$ that is true but unprovable.

\Godel{}'s sentence has a very good property, to interpret the unsolvability of it as an adversarial example.
According to Figure~\ref{figure:set}, the direct way to eliminate or alleviate adversarial examples is to shrink the adversarial set {\small$\mathds{A}(\upceil{X_\varPi,f^{X_\varPi}(X_\varPi)})$}. Therefore, we only need to expend the data set/known set {\small$\upceil{X_\varPi,f^{X_\varPi}(X_\varPi)}$}.
More technically, this solution is nothing more than adding adversarial examples, with the labels according to the desired explanation, to the data set, then training again with this new data set, known as adversarial training. We may take for granted that adversarial examples may be solved in this way, because at least these old adversarial examples should have been solved. However, this is just a misty imagination, and demon is always in the mistiness.

\Godel{}'s sentence is an adversarial example, but \Godel{}'s sentence is ineliminable. In any consistent system $T (PA \subseteq T)$, there is a \Godel{}'s sentence $\sigma_T$, which is true but unprovable in $T$, so $T$ is incomplete. If $\sigma_T$ is added to $T$ as a axiom, to form a new system $T'(T'=T \cup \{\sigma_T\})$, of course $\sigma_T$ will be both true and provable in $T'$. However, in $T'$, there will be a new \Godel{}'s sentence $\sigma_{T'}$, which is true but unprovable in $T'$, so $T'$ is still incomplete, which is guaranteed by \Godel's first incomplete theorem. This process can be repeated as any finite times as you like, but the last system you get is still incomplete, for example, the \Godel{}'s sentence of the last system is still true but unprovable. The incomplete theorem will always hold, so the adversarial example is unsolvable, because the adversarial set won't be empty forever.

\section{Incomputability for Adversarial Examples}

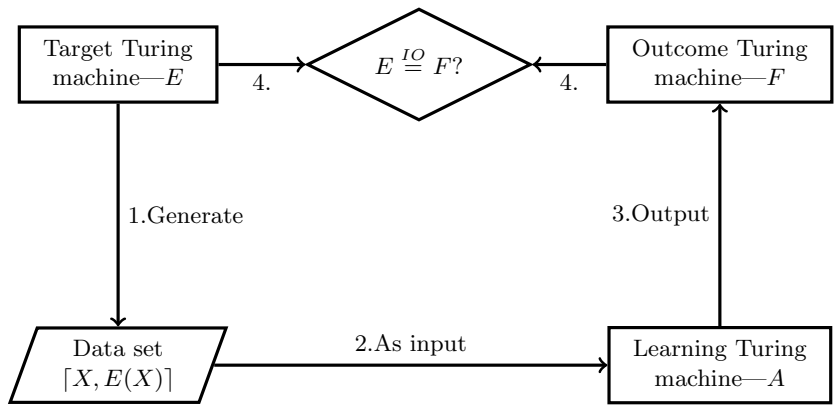
\begin{figure}[htb!]\footnotesize
\centering
\begin{tikzpicture}
[AlgoNode/.style={rectangle,draw=black!100,fill=black!0,very thick,minimum size=10mm},
DataNode/.style={trapezium, trapezium left angle=70, trapezium right angle=110, fill=black!0, draw=black!100,very thick},
AskNode/.style={diamond, aspect=2, inner sep=2pt, fill=black!0, draw=black!100,very thick}]
\node[AskNode] (ask) at(4,4){\begin{tabular}{c}
                             $E \IOeq F$?
                             \end{tabular}};
\node[DataNode] (data) at(0,0) {\begin{tabular}{c}
                                Data set\\
                                $\upceil{X,E(X)}$
                                \end{tabular}};
\node[AlgoNode] (learning) at(8,0) {\begin{tabular}{c}
                                  Learning Turing\\
                                  machine---$A$
                                  \end{tabular}};
\node[AlgoNode] (target) at(0,4) {\begin{tabular}{c}
                                Target Turing\\
                                machine---$E$
                                \end{tabular}};
\node[AlgoNode] (outcome) at(8,4) {\begin{tabular}{c}
                                 Outcome Turing\\
                                 machine---$F$
                                 \end{tabular}};
\draw[color=black!100,very thick,->] (data) -- node[above]{2.As input} (learning);
\draw[color=black!100,very thick,->] (learning) -- node[left]{3.Output} (outcome);
\draw[color=black!100,very thick,->] (target) -- node[right]{1.Generate} (data);
\draw[color=black!100,very thick,->] (target) -- node[below]{4.} (ask);
\draw[color=black!100,very thick,->] (outcome) -- node[below]{4.} (ask);
\end{tikzpicture}
\caption{Dependence in $\DesiredOne$ in Theorem~\ref{thm:DesiredOne}}
\label{figure:machinelearning}
\end{figure}

The adversarial example above is unsolvable, because there are always at least two non-isomorphic natural explanations that can explain the data set. However, that the adversarial set can't be empty doesn't mean what the learning algorithm has learned must be the undesired explanation. What if the learning algorithm happens to learn the desired one? First, Turing does not believe this telepathy deserves attention~\citep{turing2009computing}, and neither do we. Second, in Section~\ref{section:Peirce'sLaw}, we have shown that the idea of the builders of the data set doesn't matter at all. Third, taking all information into account, like the details of learning algorithm, the details of data set and the performance on training set, verification set or test set, nothing can be concluded about it. Finally, even if this lottery thing is to happen, we can't know it effectively that we are to win the lottery. Similarly, for any Turing machine and any input string, the Turing machine must halt or not halt on it, but we can't know it effectively, which is famous as the \emph{halting problem} of Turing machine, and it's undecidable\citep{turing1937computable}. We will prove that whether a learning algorithm can learn the desired one is unrecognizable (Theorem~\ref{thm:DesiredOne}). The description of symbols is in Table~\ref{table:turing} and dependence in $\DesiredOne$ in Theorem~\ref{thm:DesiredOne} is in Figure~\ref{figure:machinelearning}.
Therefore, even if the outputs for all adversarial examples indeed conform to the explanation we desire, we can't know it effectively by algorithm or computing.

\begin{table}[h]\small
\centering
\begin{tabular}{|p{3cm}|p{11cm}|}
\hline
\multicolumn{1}{|c|}{Symbol} &\multicolumn{1}{c|}{Description}\\
\hline
$M,M'$ & Turing machine.\\
\hline
$\omega,x$ & input string.\\
\hline
$\TM$ &Turing machine space, the set consisting of all Turing machines.\\
\hline
$H(M)$ & the set consisting of all strings, on which $M$ halts.\\
\hline
$M(\omega)$ &  the output of $M$ on $\omega$, $\omega \in H(M)$.\\
\hline
$\upceil{X,M(X)}$ & $\{(x,M(x)) \mid x \in X\}$\\
\hline
$\lrangle{O_1,O_2,\cdots,O_n}$ & the encoding string of objects $O_1,O_2,\cdots,O_n$.\\
\hline
$L_1 \manyone L_2$ & Language $L_1$ is many-one reducible to language $L_2$.\\
\hline
\end{tabular}
\caption{Description of symbols}
\label{table:turing}
\end{table}

\vspace{-0.2cm}

\begin{definition} \label{defn:IOeq}
For any Turing machine $M$ and $M'$, $M$ and $M'$ are input-output equivalent if and only if $H(M)=H(M')$ and $\forall \omega \in H(M),M(\omega)=M' (\omega)$, denoted as $M \IOeq M'$.
\end{definition}

\begin{theorem} \label{thm:reductionrecognizable-}
If $L_1 \manyone L_2$ and $L_1$ is unrecognizable, $L_2$ is unrecognizable.
\end{theorem}

\begin{center}
$\HALT=\{\lrangle{M,\omega} \mid M \in \TM,\omega \in H(M).\}$
\end{center}
\vspace{-0.5cm}

\begin{theorem}[Halting problem] \label{thm:HALT}
$\HALTc$ is undecidable and unrecognizable.
\end{theorem}

\begin{center}
$\SAMEIO = \{\lrangle{M_1,M_2} \mid M_1,M_2 \in \TM,M_1 \IOeq M_2.\}$
\end{center}
\vspace{-0.5cm}

\begin{theorem} \label{thm:SAMEIO}
$\SAMEIO$ is unrecognizable.
\end{theorem}
\begin{proofx}
We prove $\HALTc \manyone \SAMEIO$, so we need to prove $G: \HALT \manyone \SAMEIOc$:

$G$ = ``For input $\lrangle{M,\omega}$, in which $M \in \TM$ and $\omega$ is a string:
\begin{enumerate}\setlength{\itemindent}{4em}
  \item Construct the following two machines $M_1$ and $M_2$.
  \begin{description}\setlength{\itemindent}{1.5em}
    \item[] $M_1$ = `For any input $x$:
    \begin{enumerate}\setlength{\itemindent}{5em}
      \item Be circular.'
    \end{enumerate}
    \item[] $M_2$ = `For any input $x$:
    \begin{enumerate}\setlength{\itemindent}{5em}
      \item Run $M$ on $\omega$. If $M$ halts on $\omega$, output $\omega$ and halt.'
    \end{enumerate}
  \end{description}
  \item Output $\lrangle{M_1,M_2}$.''\hfill\BlackBox\\[2mm]
\end{enumerate}
\end{proofx}

\begin{center}
$\DesiredOne=\{\lrangle{A,E,X} \mid A,E \in \TM, X \subseteq H(E),A(\lrangle{\upceil{X,E(X)}})=\lrangle{F},E \IOeq F.\}$
\end{center}
\vspace{-0.5cm}

\begin{theorem} \label{thm:DesiredOne}
$\DesiredOne$ is unrecognizable.
\end{theorem}
\begin{proofx}
$G:\SAMEIO \manyone \DesiredOne$:

$G$ = ``For input $\lrangle{M_1,M_2}$, in which $M_1,M_2 \in \TM$:
\begin{enumerate}\setlength{\itemindent}{4em}
  \item Construct the machine $A$.
  \begin{description}\setlength{\itemindent}{1.5em}
    \item[] $A$ = `For any input $\omega$:
    \begin{enumerate}\setlength{\itemindent}{5em}
      \item Output $\lrangle{M_2}$ and halt.'
    \end{enumerate}
  \end{description}
  \item Output $\lrangle{A,M_1,\varnothing}$.''\hfill\BlackBox\\[2mm]
\end{enumerate}
\end{proofx}

\section{Conclusion}

In this paper, we show the reason for the existence of adversarial examples is there are non-isomorphic natural explanations that can all explain data set.
Specifically, G{\"o}del's sentence is an adversarial example but ineliminable, because the two natural explanations of being true and provable are always non-isomorphic, which is guaranteed by \Godel's first incomplete theorem. Therefore, it can't be solved by the re-accumulation of data set or the re-improvement of learning algorithm. Any data set can't eliminate the inherent incompleteness and any learning algorithm can't distinguish which explanation is the desired one.
Finally, we prove the incomputability for adversarial examples that whether a learning algorithm can learn the desired explanation is unrecognizable.

\vskip 0.2in
\bibliography{Ref}

\begin{thebibliography}{44}
\providecommand{\natexlab}[1]{#1}
\providecommand{\url}[1]{\texttt{#1}}
\expandafter\ifx\csname urlstyle\endcsname\relax
  \providecommand{\doi}[1]{doi: #1}\else
  \providecommand{\doi}{doi: \begingroup \urlstyle{rm}\Url}\fi

\bibitem[Abbasi and Gagn{\'e}(2017)]{abbasi2017robustness}
Mahdieh Abbasi and Christian Gagn{\'e}.
\newblock Robustness to adversarial examples through an ensemble of
  specialists.
\newblock \emph{arXiv preprint arXiv:1702.06856}, 2017.

\bibitem[Ackerman()]{TeslaAutopilotThreeSmallStickersIEEE26}
Evan Ackerman.
\newblock Three small stickers in intersection can cause tesla autopilot to
  swerve into wrong lane.
\newblock
  \url{https://spectrum.ieee.org/cars-that-think/transportation/self-driving/three-small-stickers-on-road-can-steer-tesla-autopilot-into-oncoming-lane}.
\newblock Accessed September 14, 2019.

\bibitem[Alcorn et~al.(2019)Alcorn, Li, Gong, Wang, Mai, Ku, and
  Nguyen]{alcorn2019strike}
Michael~A Alcorn, Qi~Li, Zhitao Gong, Chengfei Wang, Long Mai, Wei-Shinn Ku,
  and Anh Nguyen.
\newblock Strike (with) a pose: Neural networks are easily fooled by strange
  poses of familiar objects.
\newblock In \emph{Proceedings of the IEEE Conference on Computer Vision and
  Pattern Recognition}, pages 4845--4854, 2019.

\bibitem[Alzantot et~al.(2018)Alzantot, Sharma, Elgohary, Ho, Srivastava, and
  Chang]{alzantot2018generating}
Moustafa Alzantot, Yash Sharma, Ahmed Elgohary, Bo-Jhang Ho, Mani Srivastava,
  and Kai-Wei Chang.
\newblock Generating natural language adversarial examples.
\newblock \emph{arXiv preprint arXiv:1804.07998}, 2018.

\bibitem[Biggio et~al.(2013)Biggio, Corona, Maiorca, Nelson, {\v{S}}rndi{\'c},
  Laskov, Giacinto, and Roli]{biggio2013evasion}
Battista Biggio, Igino Corona, Davide Maiorca, Blaine Nelson, Nedim
  {\v{S}}rndi{\'c}, Pavel Laskov, Giorgio Giacinto, and Fabio Roli.
\newblock Evasion attacks against machine learning at test time.
\newblock In \emph{Joint European conference on machine learning and knowledge
  discovery in databases}, pages 387--402. Springer, 2013.

\bibitem[Bubeck et~al.(2018)Bubeck, Price, and
  Razenshteyn]{bubeck2018adversarial}
S{\'e}bastien Bubeck, Eric Price, and Ilya Razenshteyn.
\newblock Adversarial examples from computational constraints.
\newblock \emph{arXiv preprint arXiv:1805.10204}, 2018.

\bibitem[Carlini and Wagner(2018)]{carlini2018audio}
Nicholas Carlini and David Wagner.
\newblock Audio adversarial examples: Targeted attacks on speech-to-text.
\newblock In \emph{2018 IEEE Security and Privacy Workshops (SPW)}, pages 1--7.
  IEEE, 2018.

\bibitem[Crawford(2016)]{crawford2016artificial}
Kate Crawford.
\newblock Artificial intelligence’s white guy problem.
\newblock \emph{The New York Times}, 25, 2016.

\bibitem[Eger et~al.(2019)Eger, {\c{S}}ahin, R{\"u}ckl{\'e}, Lee, Schulz,
  Mesgar, Swarnkar, Simpson, and Gurevych]{eger2019text}
Steffen Eger, G{\"o}zde~G{\"u}l {\c{S}}ahin, Andreas R{\"u}ckl{\'e}, Ji-Ung
  Lee, Claudia Schulz, Mohsen Mesgar, Krishnkant Swarnkar, Edwin Simpson, and
  Iryna Gurevych.
\newblock Text processing like humans do: Visually attacking and shielding nlp
  systems.
\newblock \emph{arXiv preprint arXiv:1903.11508}, 2019.

\bibitem[Engstrom et~al.(2017)Engstrom, Tran, Tsipras, Schmidt, and
  Madry]{engstrom2017rotation}
Logan Engstrom, Brandon Tran, Dimitris Tsipras, Ludwig Schmidt, and Aleksander
  Madry.
\newblock A rotation and a translation suffice: Fooling cnns with simple
  transformations.
\newblock \emph{arXiv preprint arXiv:1712.02779}, 2017.

\bibitem[Eykholt et~al.(2018)Eykholt, Evtimov, Fernandes, Li, Rahmati, Xiao,
  Prakash, Kohno, and Song]{eykholt2018robust}
Kevin Eykholt, Ivan Evtimov, Earlence Fernandes, Bo~Li, Amir Rahmati, Chaowei
  Xiao, Atul Prakash, Tadayoshi Kohno, and Dawn Song.
\newblock Robust physical-world attacks on deep learning visual classification.
\newblock In \emph{Proceedings of the IEEE Conference on Computer Vision and
  Pattern Recognition}, pages 1625--1634, 2018.

\bibitem[Ford et~al.(2019)Ford, Gilmer, Carlini, and
  Cubuk]{ford2019adversarial}
Nic Ford, Justin Gilmer, Nicolas Carlini, and Dogus Cubuk.
\newblock Adversarial examples are a natural consequence of test error in
  noise.
\newblock \emph{arXiv preprint arXiv:1901.10513}, 2019.

\bibitem[Ghorbani et~al.(2019)Ghorbani, Wexler, Zou, and
  Kim]{ghorbani2019towards}
Amirata Ghorbani, James Wexler, James~Y Zou, and Been Kim.
\newblock Towards automatic concept-based explanations.
\newblock In \emph{Advances in Neural Information Processing Systems}, pages
  9273--9282, 2019.

\bibitem[G{\"o}del(1931)]{godel1931formal}
Kurt G{\"o}del.
\newblock {\"U}ber formal unentscheidbare s{\"a}tze der principia mathematica
  und verwandter systeme i.
\newblock \emph{Monatshefte f{\"u}r mathematik und physik}, 38\penalty0
  (1):\penalty0 173--198, 1931.

\bibitem[Goodfellow et~al.(2014)Goodfellow, Shlens, and
  Szegedy]{goodfellow2014explaining}
Ian~J Goodfellow, Jonathon Shlens, and Christian Szegedy.
\newblock Explaining and harnessing adversarial examples.
\newblock \emph{arXiv preprint arXiv:1412.6572}, 2014.

\bibitem[Gu and Rigazio(2014)]{gu2014towards}
Shixiang Gu and Luca Rigazio.
\newblock Towards deep neural network architectures robust to adversarial
  examples.
\newblock \emph{arXiv preprint arXiv:1412.5068}, 2014.

\bibitem[Hendrycks et~al.(2019)Hendrycks, Zhao, Basart, Steinhardt, and
  Song]{hendrycks2019natural}
Dan Hendrycks, Kevin Zhao, Steven Basart, Jacob Steinhardt, and Dawn Song.
\newblock Natural adversarial examples.
\newblock \emph{arXiv preprint arXiv:1907.07174}, 2019.

\bibitem[Ilyas et~al.(2019)Ilyas, Santurkar, Tsipras, Engstrom, Tran, and
  Madry]{ilyas2019adversarial}
Andrew Ilyas, Shibani Santurkar, Dimitris Tsipras, Logan Engstrom, Brandon
  Tran, and Aleksander Madry.
\newblock Adversarial examples are not bugs, they are features.
\newblock \emph{arXiv preprint arXiv:1905.02175}, 2019.

\bibitem[Katz et~al.(2017)Katz, Barrett, Dill, Julian, and
  Kochenderfer]{katz2017reluplex}
Guy Katz, Clark Barrett, David~L Dill, Kyle Julian, and Mykel~J Kochenderfer.
\newblock Reluplex: An efficient smt solver for verifying deep neural networks.
\newblock In \emph{International Conference on Computer Aided Verification},
  pages 97--117. Springer, 2017.

\bibitem[Komkov and Petiushko(2019)]{komkov2019advhat}
Stepan Komkov and Aleksandr Petiushko.
\newblock Advhat: Real-world adversarial attack on arcface face id system.
\newblock \emph{arXiv preprint arXiv:1908.08705}, 2019.

\bibitem[Lei et~al.(2019)Lei, Wu, Chen, Dimakis, Dhillon, and
  Witbrock]{lei2019discrete}
Qi~Lei, Lingfei Wu, Pin-Yu Chen, Alexandros~G Dimakis, Inderjit~S Dhillon, and
  Michael Witbrock.
\newblock Discrete adversarial attacks and submodular optimization with
  applications to text classification.
\newblock \emph{Systems and Machine Learning (SysML)}, 2019.

\bibitem[Li et~al.(2019)Li, Qu, Li, Szurley, Kolter, and
  Metze]{li2019adversarial}
Juncheng Li, Shuhui Qu, Xinjian Li, Joseph Szurley, J~Zico Kolter, and Florian
  Metze.
\newblock Adversarial music: Real world audio adversary against wake-word
  detection system.
\newblock In \emph{Advances in Neural Information Processing Systems}, pages
  11908--11918, 2019.

\bibitem[Liu et~al.(2019)Liu, Liu, Fan, Ma, Zhang, Xie, and
  Tao]{liu2019perceptual}
Aishan Liu, Xianglong Liu, Jiaxin Fan, Yuqing Ma, Anlan Zhang, Huiyuan Xie, and
  Dacheng Tao.
\newblock Perceptual-sensitive gan for generating adversarial patches.
\newblock In \emph{AAAI}, 2019.

\bibitem[Lu et~al.(2017)Lu, Issaranon, and Forsyth]{lu2017safetynet}
Jiajun Lu, Theerasit Issaranon, and David Forsyth.
\newblock Safetynet: Detecting and rejecting adversarial examples robustly.
\newblock In \emph{Proceedings of the IEEE International Conference on Computer
  Vision}, pages 446--454, 2017.

\bibitem[Nakkiran(2019)]{nakkiran2019adversarial}
Preetum Nakkiran.
\newblock Adversarial robustness may be at odds with simplicity.
\newblock \emph{arXiv preprint arXiv:1901.00532}, 2019.

\bibitem[Niven and Kao(2019)]{niven2019probing}
Timothy Niven and Hung-Yu Kao.
\newblock Probing neural network comprehension of natural language arguments.
\newblock In \emph{Proceedings of the 57th Annual Meeting of the Association
  for Computational Linguistics}, pages 4658--4664, 2019.

\bibitem[Papernot et~al.(2016)Papernot, McDaniel, Wu, Jha, and
  Swami]{papernot2016distillation}
Nicolas Papernot, Patrick McDaniel, Xi~Wu, Somesh Jha, and Ananthram Swami.
\newblock Distillation as a defense to adversarial perturbations against deep
  neural networks.
\newblock In \emph{2016 IEEE Symposium on Security and Privacy (SP)}, pages
  582--597. IEEE, 2016.

\bibitem[Papernot et~al.(2017)Papernot, McDaniel, Goodfellow, Jha, Celik, and
  Swami]{papernot2017practical}
Nicolas Papernot, Patrick McDaniel, Ian Goodfellow, Somesh Jha, Z~Berkay Celik,
  and Ananthram Swami.
\newblock Practical black-box attacks against machine learning.
\newblock In \emph{Proceedings of the 2017 ACM on Asia conference on computer
  and communications security}, pages 506--519. ACM, 2017.

\bibitem[Qin et~al.(2019)Qin, Carlini, Goodfellow, Cottrell, and
  Raffel]{qin2019imperceptible}
Yao Qin, Nicholas Carlini, Ian Goodfellow, Garrison Cottrell, and Colin Raffel.
\newblock Imperceptible, robust, and targeted adversarial examples for
  automatic speech recognition.
\newblock \emph{arXiv preprint arXiv:1903.10346}, 2019.

\bibitem[Rosser(1936)]{rosser1936extensions}
Barkley Rosser.
\newblock Extensions of some theorems of g{\"o}del and church.
\newblock \emph{The journal of symbolic logic}, 1\penalty0 (3):\penalty0
  87--91, 1936.

\bibitem[Samanta and Mehta(2017)]{samanta2017towards}
Suranjana Samanta and Sameep Mehta.
\newblock Towards crafting text adversarial samples.
\newblock \emph{arXiv preprint arXiv:1707.02812}, 2017.

\bibitem[Schmidt et~al.(2018)Schmidt, Santurkar, Tsipras, Talwar, and
  Madry]{schmidt2018adversarially}
Ludwig Schmidt, Shibani Santurkar, Dimitris Tsipras, Kunal Talwar, and
  Aleksander Madry.
\newblock Adversarially robust generalization requires more data.
\newblock In \emph{Advances in Neural Information Processing Systems}, pages
  5014--5026, 2018.

\bibitem[Shafahi et~al.(2018)Shafahi, Huang, Studer, Feizi, and
  Goldstein]{shafahi2018adversarial}
Ali Shafahi, W~Ronny Huang, Christoph Studer, Soheil Feizi, and Tom Goldstein.
\newblock Are adversarial examples inevitable?
\newblock \emph{arXiv preprint arXiv:1809.02104}, 2018.

\bibitem[Stutz et~al.(2019)Stutz, Hein, and Schiele]{stutz2019disentangling}
David Stutz, Matthias Hein, and Bernt Schiele.
\newblock Disentangling adversarial robustness and generalization.
\newblock In \emph{Proceedings of the IEEE Conference on Computer Vision and
  Pattern Recognition}, pages 6976--6987, 2019.

\bibitem[Szegedy et~al.(2013)Szegedy, Zaremba, Sutskever, Bruna, Erhan,
  Goodfellow, and Fergus]{szegedy2013intriguing}
Christian Szegedy, Wojciech Zaremba, Ilya Sutskever, Joan Bruna, Dumitru Erhan,
  Ian Goodfellow, and Rob Fergus.
\newblock Intriguing properties of neural networks.
\newblock \emph{arXiv preprint arXiv:1312.6199}, 2013.

\bibitem[Taori et~al.(2019)Taori, Kamsetty, Chu, and Vemuri]{taori2019targeted}
Rohan Taori, Amog Kamsetty, Brenton Chu, and Nikita Vemuri.
\newblock Targeted adversarial examples for black box audio systems.
\newblock In \emph{2019 IEEE Security and Privacy Workshops (SPW)}, pages
  15--20. IEEE, 2019.

\bibitem[Thys et~al.(2019)Thys, Van~Ranst, and Goedem{\'e}]{thys2019fooling}
Simen Thys, Wiebe Van~Ranst, and Toon Goedem{\'e}.
\newblock Fooling automated surveillance cameras: adversarial patches to attack
  person detection.
\newblock In \emph{Proceedings of the IEEE Conference on Computer Vision and
  Pattern Recognition Workshops}, pages 0--0, 2019.

\bibitem[Turing(1937)]{turing1937computable}
Alan~M Turing.
\newblock On computable numbers, with an application to the
  entscheidungsproblem.
\newblock \emph{Proceedings of the London mathematical society}, 2\penalty0
  (1):\penalty0 230--265, 1937.

\bibitem[Turing(2009)]{turing2009computing}
Alan~M Turing.
\newblock Computing machinery and intelligence.
\newblock In \emph{Parsing the Turing Test}, pages 23--65. Springer, 2009.

\bibitem[Weiguang~Ding et~al.(2019)Weiguang~Ding, Yik Chau~Lui, Jin, Wang, and
  Huang]{weiguang2019sensitivity}
Gavin Weiguang~Ding, Kry Yik Chau~Lui, Xiaomeng Jin, Luyu Wang, and Ruitong
  Huang.
\newblock On the sensitivity of adversarial robustness to input data
  distributions.
\newblock In \emph{Proceedings of the IEEE Conference on Computer Vision and
  Pattern Recognition Workshops}, pages 13--16, 2019.

\bibitem[Wu et~al.(2017)Wu, Bamman, and Russell]{wu2017adversarial}
Yi~Wu, David Bamman, and Stuart Russell.
\newblock Adversarial training for relation extraction.
\newblock In \emph{Proceedings of the 2017 Conference on Empirical Methods in
  Natural Language Processing}, pages 1778--1783, 2017.

\bibitem[Xiao et~al.(2018)Xiao, Tjeng, Shafiullah, and Madry]{xiao2018training}
Kai~Y Xiao, Vincent Tjeng, Nur~Muhammad Shafiullah, and Aleksander Madry.
\newblock Training for faster adversarial robustness verification via inducing
  relu stability.
\newblock \emph{arXiv preprint arXiv:1809.03008}, 2018.

\bibitem[Yuan et~al.(2019)Yuan, He, Zhu, and Li]{yuan2019adversarial}
Xiaoyong Yuan, Pan He, Qile Zhu, and Xiaolin Li.
\newblock Adversarial examples: Attacks and defenses for deep learning.
\newblock \emph{IEEE transactions on neural networks and learning systems},
  2019.

\bibitem[Zhang et~al.(2019)Zhang, Sheng, and Alhazmi]{zhang2019generating}
Wei~Emma Zhang, Quan~Z Sheng, and Ahoud Abdulrahmn~F Alhazmi.
\newblock Adversarial attacks on deep learning models in natural language
  processing: A survey.
\newblock \emph{arXiv preprint arXiv:1901.06796}, 2019.

\end{thebibliography}

\end{document}